\documentclass[lettersize,journal]{IEEEtran}
\usepackage{amsmath,amsfonts}
\usepackage{algorithmic}
\usepackage{algorithm}
\usepackage{array}
\usepackage[caption=false,font=normalsize,labelfont=sf,textfont=sf]{subfig}
\usepackage{textcomp}
\usepackage{stfloats}
\usepackage{url}
\usepackage{verbatim}
\usepackage{graphicx}
\usepackage{cite}
\usepackage{color}
\usepackage{booktabs}   
\usepackage{multirow}   
\usepackage{graphicx}   
\usepackage{arydshln}   
\usepackage[table]{xcolor} 
\usepackage{xspace}

\newcommand{\ourrow}{\rowcolor{gray!10}}
\def\pilot{\textbf{\textsc{PILOT}}\xspace}
\begin{document}

\title{\textbf{\textsc{PILOT}}: A Perceptive Integrated Low-level Controller for Loco-manipulation over Unstructured Scenes}

\author{Xinru Cui, Linxi Feng, Yixuan Zhou, Haoqi Han, Zhe Liu, and  Hesheng Wang
\vspace{-20pt}
\thanks{

Xinru Cui, Linxi Feng and Zhe Liu are with the School of Automation and Intelligent Sensing, Shanghai Jiao Tong University, Shanghai, 200240, China. 
Yixuan Zhou is with Global College, Shanghai Jiao Tong University, Shanghai, 200240, China. Haoqi Han is with the school of Computer Science, Shanghai Jiao Tong University, Shanghai, 200240, China.

Hesheng Wang is with the School of Automation and Intelligent Sensing,
and the Shanghai Key Laboratory of Navigation and Location Based Services, Shanghai Jiao Tong University, Shanghai 200240, China.

Corresponding Author: Hesheng Wang (email:wanghesheng@sjtu.edu.cn)
}
}

\markboth{Journal of \LaTeX\ Class Files,~Vol.~14, No.~8, August~2021}%
{Shell \MakeLowercase{\textit{et al.}}: A Sample Article Using IEEEtran.cls for IEEE Journals}


\maketitle

\begin{abstract}
Humanoid robots hold great potential for diverse interactions and daily service tasks within human-centered environments, necessitating controllers that seamlessly integrate precise locomotion with dexterous manipulation.
However, most existing whole-body controllers lack exteroceptive awareness of the surrounding environment, rendering them insufficient for stable task execution in complex, unstructured scenarios.
To address this challenge, we propose \pilot, a unified single-stage reinforcement learning (RL) framework tailored for perceptive loco-manipulation, which synergizes perceptive locomotion and expansive whole-body control within a single policy.
To enhance terrain awareness and ensure precise foot placement, we design a cross-modal context encoder that fuses prediction-based proprioceptive features with attention-based perceptive representations. Furthermore, we introduce a Mixture-of-Experts (MoE) policy architecture to coordinate diverse motor skills, facilitating better specialization across distinct motion patterns. Extensive experiments in both simulation and on the physical Unitree G1 humanoid robot validate the efficacy of our framework. 
\pilot demonstrates superior stability, command tracking precision, and terrain traversability compared to existing baselines. These results highlight its potential to serve as a robust, foundational low-level controller for loco-manipulation in unstructured scenes.
\end{abstract}
\vspace{-10pt}

\section{Introduction}
\label{sec1}
\IEEEPARstart{H}{umans} substantially expand their reachable workspace through coordinated whole-body movements and by traversing complex, unstructured environments. For example, a person can squat and bend to pick a box from the floor and then carry it across stairs or steps. Humanoid robots, by virtue of their human-like morphology, are intrinsically well suited for loco-manipulation in such unstructured settings. This potential paves the way for humanoid robots to be seamlessly integrated into human life and to effectively assist with daily tasks, a goal that hinges upon robust locomotion, precise manipulation, and coordinated whole-body movements.

Traditional model-based methods typically rely on accurate system identification and precise dynamic models \cite{model_based1,model_based2,model_based3}. As a consequence, they struggle with unmodeled disturbances and environmental variability, which makes whole-body control in complex, unknown scenes difficult to deploy robustly. By contrast, end-to-end learning methods based on reinforcement learning have recently made substantial progress in humanoid whole-body control \cite{learning1,learning2,learning3,learning4,learning5,learning6,PIM}, laying a foundation for learning-based loco-manipulation. Techniques such as teacher–student architectures \cite{TWIST}, adaptive motion optimization \cite{AMO}, and sequential skill acquisition \cite{ULC} have enabled robots to reach further, manipulate more reliably, and execute richer interactions in structured environments. Nonetheless, two critical challenges remain unresolved and together constitute a major barrier to deploying humanoids in unstructured environments.

The first challenge centers on traversal-aware loco-manipulation over non-planar scenes. 
While the bipedal morphology inherently affords superior traversability and reachability compared to wheeled platforms, recent low-level whole-body controllers remain largely confined to planar or mildly varying surfaces \cite{HOMIE,ULC,CLONE,AMO}.
In such cases, the control policies operate in a blind fashion, producing locomotion and manipulation commands without explicit awareness of underlying geometric variations in the environment. While sufficient for flat-ground operation, it fundamentally limits applicability when humanoid robots are required to traverse stairs or rugged terrain while concurrently executing manipulation tasks. Under teleoperation, a blind controller forces the human operator to compensate for missing terrain adaptation, increasing cognitive load and safety risk. Under autonomous execution, blind control becomes even more critical: without terrain-aware adaptation, the robot cannot ensure balance or safety during elevation transitions, and task-level strategies that rely on such controllers are inherently fragile. Addressing this challenge requires tightly coupling perception of terrain geometry, temporally consistent foothold reasoning, and whole-body adjustments so that mobility and manipulation remain safe and effective across extended, non-planar traversals.
 
The second challenge concerns multi-task, coordinated whole-body control over large workspaces. Some works separate lower-body and upper-body control to simplify design and improve stability \cite{HOMIE,THOR}, but this decomposition sacrifices the natural synergies necessary for fluid whole-body behaviors. A single unified policy that concurrently controls legs, torso, and arms—and that must coordinate actions across large reachable workspaces—promises more capable and expressive behaviors, but it faces severe technical obstacles: very high-dimensional observation and action spaces, and heterogeneous, often conflicting objectives (e.g., balance and robustness versus end-effector precision over large reaches). To bootstrap learning, some works incorporate motion-capture data as tracking references \cite{CLONE,TWIST}; yet such trajectories are typically only kinematically feasible, fail to account for the robot’s dynamics, and introduce distribution bias into the training data, which can produce fragile or unsafe behavior under out-of-distribution deployment. These difficulties are further compounded when whole-body control is unified with visual perception into a single policy: The inclusion of terrain awareness increases input dimensionality and perceptual uncertainty in the multi-task problem, which may introduce conflicting gradient signals among objectives, thereby complicating stable learning and reliable generalization.

To address the above challenges, we propose \pilot, a perceptive and unified loco-manipulation controller designed for unstructured scenes. We incorporate a robot-centric, LiDAR-based elevation map to capture surrounding terrain information, ensuring target reachability during three-dimensional loco-manipulation. Recognizing that precise foot placement and the minimization of horizontal foot-terrain collisions are critical for the stability of upper-limb manipulation during loco-manipulation tasks, we employ an attention-based multi-scale perception fusion and encoding architecture to enhance environmental understanding. We also utilize principled and progressive random command sampling rather than relying on motion capture data to mitigate distribution bias and comprehensively cover the feasible command space. Finally, we leverage a Mixture-of-Experts (MoE) architecture to coordinate diverse motor skills into a unified control policy.

Building upon this perceptive and generalizable controller, we utilize a lightweight VR interface, comprising a headset and handheld controllers, to teleoperate the robot in executing long-horizon loco-manipulation tasks while traversing complex terrains like stairs and high steps. Moreover, we present a hierarchical reinforcement learning strategy that empowers the robot to autonomously perform loco-manipulation tasks, 
such as navigating to a target location and bending to retrieve a box. 
Extensive experimental evaluations in both simulation and the real world demonstrate that our method achieves superior performance across various loco-manipulation tasks on complex terrains, highlighting its versatility and robustness.

To summarize, our main contributions are as follows:
\begin{itemize}
    \item We propose \pilot, a unified one-stage RL framework that seamlessly integrates perceptive locomotion with large-workspace whole-body control. This framework serves as a robust low-level controller, enabling humanoid loco-manipulation across complex terrains.
    \item We design an attention-based multi-scale perception encoder to enhance terrain traversal stability through precise foot placement. Furthermore, we incorporate a MoE architecture to facilitate the unified learning of diverse motor skills, ensuring natural and coordinated transitions between locomotion and manipulation behaviors.
    \item We conduct extensive real-world experiments on the Unitree G1 humanoid robot using both teleoperation and autonomous policies. The results demonstrate the effectiveness of our approach in achieving robust loco-manipulation in three-dimensional complex scenes.
\end{itemize}

\section{Related Works}
\subsection{Learning Humanoid Locomotion}
The control of legged locomotion has witnessed a paradigm shift from model-based optimization \cite{model_based1,model_based2,model_based3 } to data-driven reinforcement learning \cite{learning1,learning2,learning3,learning4,learning5,learning6,PIM}. Early learning-based approaches primarily focused on blind locomotion, utilizing proprioceptive feedback to robustly traverse flat ground and handle perturbations \cite{learning2,learning3,blind}.
However, blind policies fundamentally struggle in unstructured environments characterized by discrete obstacles or significant height variations, as they lack predictive terrain awareness. To address this, recent research has integrated exteroceptive perception into the control loop. A common approach involves coupling RL with Elevation Maps \cite{elevation1,elevation2}. For instance, PIM \cite{PIM} integrates a LiDAR-based elevation map with a hybrid internal model to enhance state estimation, enabling humanoids to traverse stairs and gaps with high success rates. He et al. \cite{learning6} employ a convolutional neural network (CNN) to embed point-wise local
terrain features in the elevation map and train an attention-based map encoding to achieve precise and generalized locomotion on sparse terrains. Gallant \cite{gallant} proposes a voxel-grid-based framework, utilizing z-grouped 2D CNNs to process full 3D terrain information for navigation. 

Despite these advancements in mobility, a critical limitation persists: most locomotion-centric policies treat the upper body essentially as a passive payload for balance maintenance or angular momentum regulation. They lack the capability to actively coordinate upper-body joints for precise manipulation tasks, limiting their applicability in real-world scenarios that require simultaneous mobility and interaction. To this end, \pilot unifies perceptive locomotion with large-workspace whole-body control within a single policy, establishing a perceptive and robust low-level controller for complex scenes.

\subsection{Humanoid Loco-Manipulation Controller}

Achieving coordinated loco-manipulation presents the challenge of managing a high-dimensional state-action space and the dynamic coupling between the floating base and end-effectors. Existing approaches generally fall into two categories: Decoupled and Unified architectures. Decoupled methods mitigate the curse of dimensionality by separating lower-body locomotion from upper-body manipulation. For instance, HOMIE \cite{HOMIE} employs RL for robust leg control while using a PD controller for the arms and waist yaw, which simplifies training but results in rigid upper-body behavior and a restricted workspace. FALCON \cite{FALCON} utilizes a dual-agent RL framework with a force curriculum to perform forceful loco-manipulation.
Hybrid approaches like AMO \cite{AMO} integrate RL-based locomotion with trajectory optimization for adaptive whole-body control, where the optimizer generates lower-body reference motions based on upper-body goals. While decoupled methods are training-efficient, they often sacrifice coordination and yield suboptimal whole-body dynamics, as the subsystems may generate conflicting control objectives.

\begin{figure*}[!t]
\centering

\includegraphics[width=0.98\linewidth]{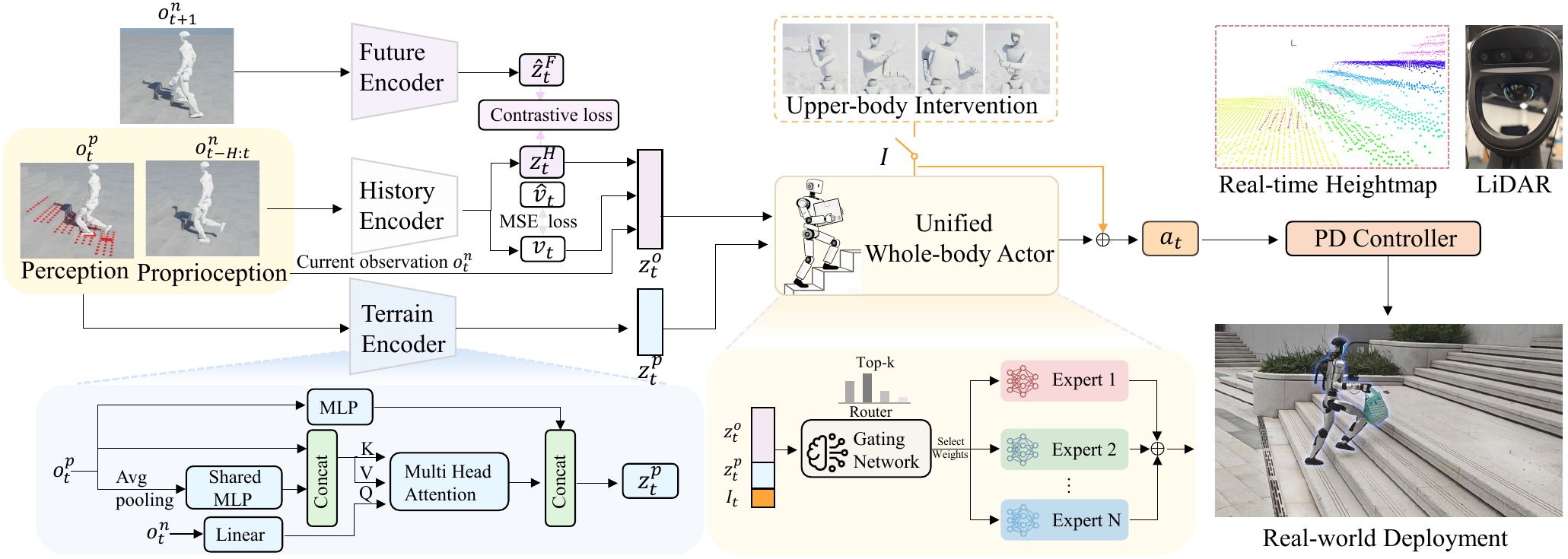}
\vspace{-8pt}
\caption{\textbf{Method overview of \pilot.} We propose a unified single-stage reinforcement learning framework that seamlessly integrates perceptive locomotion and adaptive whole-body control with an expanded workspace into a single policy. Within this framework, we leverage cross-modal context embeddings—fusing prediction-based proprioceptive features with multi-scale attention-based perceptive representations—to enhance the stability of terrain traversal. These embeddings are then fed into a MoE-based unified actor network to coordinate distinct motion skills. The resulting action outputs are converted into motor torques via a PD controller. For real-world deployment, a LiDAR-based robot-centric elevation map is utilized to capture local terrain geometry. }
\label{fig_method}
\vspace{-18pt}
\end{figure*}
In contrast, unified methods aim to control all DoFs with a single policy to maximize coordination. Some works \cite{TWIST,TWIST2,CLONE} learn whole-body control by mimicking large-scale motion capture datasets; however, these methods often suffer from inherent kinematic bias present in the training distribution. ULC \cite{ULC} proposes a unified controller trained end-to-end for multi-task tracking via careful command design and sequential skill acquisition. Nevertheless, training a monolithic unified policy remains challenging due to gradient interference between conflicting objectives and exploration inefficiency in high-dimensional spaces. Crucially, most existing loco-manipulation controllers lack exteroceptive perception capabilities, thereby limiting their applicability to flat or simple terrains. To address these limitations, we introduce an exteroceptive perception module and design a MoE architecture for whole-body coordination. This approach effectively mitigates task interference while establishing a unified, perceptive, and robust framework for complex loco-manipulation.
\section{METHOD}
In loco-manipulation tasks within complex environments, a critical challenge for humanoids lies in overcoming high-dimensional control dynamics while  ensuring locomotion stability on diverse terrains and maintaining an expansive operational workspace. To address this, \pilot is designed to learn a perceptive and unified policy, serving as a robust and versatile low-level whole-body controller for loco-manipulation.

\vspace{-5pt}
\subsection{Problem Definition}

We formulate the humanoid loco-manipulation problem as a goal-conditioned Markov Decision Process (MDP), defined by the tuple $\mathcal{M} = (\mathcal{S}, \mathcal{A}, \mathcal{G}, \mathcal{P}, \mathcal{R}, \gamma)$. Here, $\mathcal{S}$, $\mathcal{A}$, and $\mathcal{G}$ denote the state, action, and goal spaces, respectively; $\mathcal{P}$ represents the transition dynamics, $\mathcal{R}$ is the reward function, and $\gamma \in [0, 1)$ is the discount factor. At each time step $t$, the agent receives the state $s_t$ and the goal command $c_t$ and obtains a reward $r_t = \mathcal{R}(s_t, c_t)$ for policy optimization. The policy outputs an action $a_t \in \mathbb{R}^{29}$, which is subsequently fed into a PD controller to generate the desired joint torques. Finally, the Proximal Policy Optimization (PPO) algorithm is employed to maximize the expected discounted return $J=\mathbb{E}[\sum_{t=0}^{T-1}\gamma^tr_t]$.

\subsubsection{State Space Design}
The state space $\mathcal{S}$ consists of two distinct components: the proprioceptive observation $o_t^n$, which reflects the robot's internal state, and the perceptive observation $o_t^p$, which captures the surrounding terrain information. Specifically, the proprioceptive observation is defined as $o_t^n = [q_t, \dot{q}_t, \omega_t^{\text{base}}, g_t, c_t, a_{t-1}]$, which contains joint positions $q_t \in \mathbb{R}^{29}$, joint velocities $\dot{q}_t \in \mathbb{R}^{29}$, base angular velocity $\omega_t^{\text{base}} \in \mathbb{R}^3$, projected gravity $g_t \in \mathbb{R}^3$, current commands $c_t$, and previous actions $a_{t-1} \in \mathbb{R}^{29}$.
The perception $o_t^p \in \mathbb{R}^{121}$ corresponds to an egocentric elevation map of the robot's surroundings. This map is sampled on an $11 \times 11$ grid with a resolution of $0.1\,\text{m}$, where each point represents the vertical distance between the terrain and the base link. To facilitate smooth control and handle partial observability, we incorporate a multi-step history of proprioceptive observations. Consequently, the final policy observation is defined as $s_t = [o_{t-H}^n, \dots, o_{t-1}^{n},o_t^n, o_t^p]$, where $H=5$ is the history length.

\subsubsection{Goal Space Design}
The primary objective of \pilot is to achieve coordinated whole-body control that maximizes operational workspace coverage without compromising the robot's locomotion adaptability on complex terrains. To foster this capability, the robot is trained across a diverse set of challenging environments—including ascending and descending stairs, slopes, elevated platforms, and randomized rough terrains—thereby ensuring robustness in real-world loco-manipulation scenarios. Simultaneously, the robot is required to track a structured, procedurally generated command vector defined as $c_t = [v_t^x, v_t^y, \omega_t^{\text{yaw}}, h_t^{\text{base}}, \mathbf{rpy}_t, q_t^{\text{upper}^*}]$. These components specify the desired base linear velocities, base angular velocities, base height, and torso orientation, while the final term $q_t^{\text{upper}^*}$ represents the manipulation goals, denoted by the target joint configurations for the upper body.

\vspace{-8pt}
\subsection{Policy Learning}
\pilot addresses the two fundamental challenges outlined in Section \ref{sec1} through two complementary architectural components, as illustrated in Fig \ref{fig_method}. First, we introduce a cross-modal context encoder designed to effectively capture latent features from both proprioceptive and exteroceptive states. Specifically, a prediction-based proprioceptive encoder is employed to bolster state estimation, while an attention-based exteroceptive encoder facilitates robust terrain traversability and precise foothold placement. Second, we leverage a MoE-based policy network to generate coordinated whole-body actions, enabling context-aware skill adaptation and seamless transitions. These components are synergistically integrated into a unified policy framework and trained end-to-end, significantly enhancing whole-body coordination and behavioral expressiveness.

\subsubsection{Cross-modal Context Encoder} \mbox{} \\[-0.em]
\indent \textbf{Prediction-based Proprioceptive Encoder.}
Building upon the framework in \cite{PIM}, our proprioceptive history encoder is designed to implicitly model system dynamics from historical observations, thereby providing precise state estimation. The history encoder accepts a history of proprioceptive states $o_{t-H:t}^n$, alongside the current perceptive observation $o_t^p$. It is tasked with predicting the base linear velocity $\hat{v}_t$ and a latent representation of the future state $\hat{z}_{t+1}$. The velocity prediction head is supervised via a MSE loss against the ground-truth velocity obtained from the simulation. Simultaneously, the future state prediction is optimized through contrastive learning, enforcing alignment with the output of a target future encoder.

\indent \textbf{Attention-based Exteroceptive Encoder.}
Conventional approaches indiscriminately feed high-dimensional raw elevation maps into flat MLPs, relying on the policy to implicitly deduce geometric features through inefficient trial-and-error. This lack of explicit structural modeling often leads to sample inefficiency and poor adaptability to irregular terrains. To address this, we propose an attention-based multi-scale perception encoder explicitly designed to identify optimal steppable areas, thereby minimizing the risk of foot-terrain collisions (stumbles). The architecture operates on two scales: First, a global semantic feature $\phi(o_t^p)$ is extracted via an MLP to provide coarse-grained context. Second, to capture fine-grained local geometry, we employ a PointNet-inspired architecture. The elevation map $o_t^p$ undergoes average pooling followed by a shared MLP to aggregate neighborhood features, which are then concatenated with the raw map scan to yield dense point-wise local features. 

Inspired by He et al.\cite{learning6}, we implement a proprioception-guided cross-attention mechanism to extract state-dependent geometric features. In this formulation, the current proprioceptive state $o_t^n$ serves as the Query, while the generated local terrain features function as Keys and Values. This process generates a state-dependent terrain encoding, dynamically attending to critical patches—such as potential footholds or obstacles—based on the robot's current pose. The final multi-scale perceptual latent $z_t^p$ is formed by fusing this attention-weighted, fine-grained encoding with the global context $\phi(o_t^p) $:
\vspace{-5pt}
\begin{equation}
    \mathbf{z}_t^p = \left[ \phi_1({o}_t^p) \mathop{\Vert} \text{MHA}\left( \phi_2({o}_t^n), \mathcal{F}_{\text{local}}, \mathcal{F}_{\text{local}} \right) \right],
    \label{eq:attention_fusion}
\end{equation}

\vspace{-5pt}
where $[\cdot \mathop{\Vert} \cdot]$ denotes feature concatenation and $\phi_{{1,2}}$ are projection MLPs. The local feature representation is defined as $\mathcal{F}_{\text{local}} = [\psi({o}_t^p) \mathop{\Vert} {o}_t^p]$, where $\psi(\cdot)$ denotes the PointNet-inspired layer comprising average pooling and shared MLPs. MHA refers to the Multi-Head Attention mechanism \cite{attention}.
\subsubsection{Unified Whole-Body Motion Generation}\mbox{} \\[-0.em]
\indent Unlike conventional decoupled controllers that treat the upper and lower limbs separately, a unified whole-body control architecture facilitates simultaneous coordination of all degrees of freedom. This holistic approach maximizes workspace coverage and ensures tight coupling between locomotion and manipulation. However, training such unified policies is often hindered by the exploration inefficiency inherent to high-dimensional action spaces.

To mitigate this challenge, we introduce a MoE-based actor policy, designed to organically integrate diverse motor skills into a cohesive policy, enabling seamless behavioral transitions.
Structurally, the policy network comprises a gating network and a set of $N$ expert networks with distinct learnable parameters. Both the experts and the gating network receive the same input $\mathcal{S}_{input} = \{z_t^p, z_t^o,I_t\}$.
Specifically, $z_t^p$ is the perceptual latent, while $z_t^o$ serves as the proprioceptive encoding, which aggregates the history-encoded latent $z_t^H$, the estimated base velocity $v_t$, and the raw proprioceptive state $o_t^n$.
The binary indicator $I_t \in \{0, 1\}$ serves to modulate the control strategy: $I_t=0$ fixes the arm reference commands to a nominal configuration to prioritize natural, whole-body coordinated locomotion; conversely, $I_t=1$ activates a tracking mode where the policy must precisely follow sampled upper-body reference commands while ensuring lower-limb stability.

Then the gating network dynamically allocates a probability distribution across the experts, while each expert network independently proposes a candidate action distribution. To encourage specialized motion patterns, the final control action $a_t$ is synthesized via a weighted summation of the experts' outputs: $a_t = \sum_{i=1}^{N} p_i^t a_i^t$, where $p_i^t$ denotes the activation probability for the $i$-th expert, $a_i^t$ is the corresponding expert action, and $N=4$ is the number of experts.

Rather than burdening the policy with the complexity of directly regressing absolute arm joint positions from abstract commands, we adopt a residual action parameterization strategy. 
Formally, the upper-body component of the unified policy's output, denoted as $a_t^{\text{upper}}$, functions as a residual delta. It is added to the user-specified target joint configurations $q_t^{\text{upper}^*}$ to generate the final reference positions for the PD controller. This residual formulation effectively shifts the learning objective: instead of learning the full kinematics from scratch, the policy focuses exclusively on generating fine-grained corrective adjustments to compensate for dynamic disturbances and maintain whole-body balance during manipulation.

\begin{table}[tbp]
\centering
\caption{Reward Designs}
\vspace{-3pt}
\renewcommand{\arraystretch}{1.2}
\resizebox{0.98\linewidth}{!}{%
\begin{tabular}{ c  c  c  c}
\hline
\textbf{Term} & \textbf{Weight} & \textbf{Term} & \textbf{Weight} \\ \hline
\multicolumn{4}{c}{\textbf{Task Reward}} \\ \hline
Linear velocity tracking & $1.5$ &  Angular velocity tracking & $2.0$ \\
Base height tracking & $1.0$ & Torso orientation tracking & $1.0$ \\
Upper body position tracking & $1.0$ & Termination & $-200.0$\\
\hline
\multicolumn{4}{c}{\textbf{Regularization}} \\ \hline
Action rate & $-0.01$  & Energy cost & $-0.001$  \\
Joint deviation & $-0.2$ & Feet stumble & $-1.0$ \\
Feet air time & $2.0$ & Feet slippage & $-1.0$ \\
Large feet force & $-0.005$ & No fly & 0.75 \\ \hline 
\end{tabular}
}
\label{tab:reward}
\vspace{-18pt}
\end{table}

\subsubsection{Reward Design}\mbox{} \\[-0.em]
\indent As detailed in Table \ref{tab:reward}, we formulate the reward $r_t$ as a weighted sum of two distinct components: task rewards, which encourage precise tracking of the goal commands, and regularization rewards, designed to penalize unsmooth motions and mitigate physically infeasible or dangerous behaviors.

\subsubsection{Adaptive Command Curriculum and Training Strategy}\mbox{} \\[-0.em]
\indent To facilitate the progressive and robust acquisition of diverse motor skills, we implement an adaptive command curriculum. This mechanism initializes target commands within a constrained local neighborhood and dynamically expands to encompass the full command manifold as training progresses.

For the base height command, we employ a curriculum-modulated uniform sampling strategy defined as $\mathcal{U}(h_{\text{max}} - \alpha_1 (h_{\text{max}} - h_{\text{min}}), h_{\text{max}})$, where $\alpha_1 \in [0, 1]$ serves as an adaptive curriculum factor. Consequently, the task evolves from maintaining a static standing height ($h_{\text{max}}=0.76\,\text{m}$) to traversing the full vertical workspace ($[0.3, 0.76]\,\text{m}$). A similar strategy is applied to the torso orientation commands, expanding from the nominal upright pose to the complete orientation range.

Regarding upper-body manipulation, we adopt an exponential distribution sampling strategy inspired by \cite{HOMIE} to encourage smooth tracking behaviors. The sampling radius $r_{\text{upper}}$ is governed by:\\[-2em]

{\small
\begin{equation}
    r_{\text{upper}} = -\frac{1}{20(1-0.99\alpha_2)} \ln\left(1 - \mathcal{U}(0,1) \left(1 - e^{-20(1-0.99\alpha_2)}\right)\right).
    \label{eq:r_upper}
\end{equation}
}

The final target arm joint configurations are computed as:
\begin{equation}
    q^{\text{upper}^*} = q_{\text{bound}} \cdot \mathcal{U}(0, r_{\text{upper}}) \cdot \text{sign}(\mathcal{N}(0,1)),
\end{equation}
where $q_{\text{bound}}$ represents the joint limits. This strategy ensures a smooth transition from conservative, small-scale motions ($\alpha_2 \to 0$) to a comprehensive exploration of the entire joint configuration space ($\alpha_2 \to 1$).

Finally, to mitigate catastrophic forgetting and enhance sample efficiency, we design a structured training curriculum. Initially, the unified policy is trained exclusively for robust locomotion to master terrain traversability. Upon achieving a velocity tracking performance threshold, we activate randomized base height goals. Once stable height tracking is established, we sequentially introduce torso orientation targets (roll, pitch, yaw) and finally integrate the upper-body manipulation tasks using the aforementioned sampling strategy.

\section{Experiment}
In this section, we conduct comprehensive experiments in both simulations and the real world to evaluate the performance of \pilot in complex loco-manipulation tasks. The experimental validation is performed using the Unitree G1 humanoid robot, which possesses 29 degrees of freedom (DoFs). Simulation training is based on the IsaacLab simulator \cite{isaaclab} on a single NVIDIA RTX 4090 GPU. Specifically, our evaluation aims to address the following primary questions:

\textbf{Q1}: Can \pilot outperform baselines in terms of whole-body coordination and precise tracking of task commands?

\textbf{Q2}: What are the individual contributions of the attention-based perception encoder and the unified MoE policy network to \pilot's overall efficacy? 

\textbf{Q3}: How well does \pilot perform in real-world scenes?
\vspace{-20pt}

\subsection{Simulation Experiments}
\begin{table*}[t]
    \centering
    \caption{Simulation Evaluation}
      \vspace{-5pt}
  \renewcommand{\arraystretch}{0.45}
   \resizebox{0.92\linewidth}{!}{
    \begin{tabular}{@{}lcccccccl@{}}
        \toprule
        \textbf{Method}  & $E_{\mathbf{v}}$$\downarrow$ & $E_{\mathbf{\omega}}$$\downarrow$ & $E_h$$\downarrow$ & $E_r$$\downarrow$ & $E_p$$\downarrow$ & $E_y$ $\downarrow$&$E_{\mathbf{arm}}$$\downarrow$ &$E_{\mathbf{stumble}}$$\downarrow$\\

        \midrule
  \ourrow \multicolumn{9}{l}{\textbf{(a) Comparison with baselines (simple terrains)}} \\
      
  \cdashline{1-9}\noalign{\vskip 0.6mm}
  
        HOMIE & 0.386 & 0.384 & 0.022 & - & - & \textbf{0.020} & {0.340}& \quad - \\ 
        FALCON & 0.272 & 0.263 & 0.083 & - & - & 0.143 & 0.305 & \quad -\\ 
        AMO & 0.357 & - & 0.032 & 0.089 & 0.156 & 0.115 & \textbf{0.206}& \quad -\\ 
        \pilot w/o vision & \textbf{0.145} & \textbf{0.099} & \textbf{0.010} & \textbf{0.057} & \textbf{0.053} & {0.055} &  \textbf{0.206}& \quad - \\
        \midrule
      \ourrow \multicolumn{9}{l}{\textbf{(b) Ablation (full terrains)}} \\

  \cdashline{1-9}\noalign{\vskip 0.6mm}
  \pilot w/o vision& {0.201} & {0.137} & {0.013} & {0.062} &{0.070} &{0.071} &  {0.225}& 0.087\\
   \pilot w/o attention-based encoder& {0.167} & {0.128} & {0.016} & {0.065} &{0.075} &{0.073} &  {0.223}& 0.066\\
     \pilot  w/o MoE& {0.179} & {0.121} & {0.015} &{0.067} &{0.080} & {0.078} &  {0.246}& 0.017 \\
      \pilot& \textbf{0.148} & \textbf{0.102} & \textbf{0.009} & \textbf{0.055} & \textbf{0.056} & \textbf{0.068} &  \textbf{0.218}& \textbf{0.006} \\
        
        \bottomrule
       \vspace{-3pt}
    \end{tabular}
    }
    \begin{minipage}{0.9\linewidth} 
        \footnotesize 
        \textit{Note:} Baselines are excluded from full terrain evaluations due to their inability to traverse complex terrains caused by the absence of perception. We thus benchmark them against \textbf{PILOT} only on simple terrains in (a) for fairness, while (b) evaluates the proposed framework on full terrains.
    \end{minipage}
    \label{tab:errors_comparison}
    \vspace{-15pt}
\end{table*}
To answer \textbf{Q1}, we benchmark our proposed method against the following state-of-the-art baselines:
\begin{itemize}
    \item \textbf{HOMIE}\cite{HOMIE}: A decoupled controller in which locomotion is governed by an RL policy, while the waist yaw and arm joints are tracked via a PD controller.
    \item \textbf{FALCON}\cite{FALCON}: A decoupled framework employing a dual-agent learning scheme to control whole-body joints, facilitated by an adaptive force curriculum.
    \item \textbf{AMO}\cite{AMO}: A hierarchical controller that integrates RL with trajectory optimization for adaptive whole-body control. The legs and waist are managed by a lower-level policy, while the arms are controlled via PD tracking.
\end{itemize}

For quantitative comparison, we evaluate the averaged episodic command tracking errors, specifically: linear velocity ($E_\mathbf{v}$), angular velocity ($E_\omega$), base height ($E_h$), torso orientation (roll $E_r$, pitch $E_p$, yaw $E_y$), and arm joint positions ($E_{\mathbf{arm}}$). These metrics measure the discrepancy between the actual robot states and the target commands using the $L_1$ norm.

\subsubsection{\textbf{Comparison with baselines}}


It is important to note that, to the best of our knowledge, existing low-level controllers for loco-manipulation are predominantly blind, and there is a lack of established perceptive baselines specifically tailored for this domain. To ensure a fair comparison, we benchmark \textbf{a blind variant of} \pilot (with visual inputs masked) against these state-of-the-art blind baselines on simple terrains (i.e., flat ground and minor irregularities). This comparison allows us to evaluate our architecture against the current gold standard in scenarios where perception is less critical, ensuring that \pilot maintains robust loco-manipulation capabilities even in the absence of visual feedback.
As shown in Table \ref{tab:errors_comparison}(a), \pilot achieves the lowest overall tracking errors across both lower-body locomotion and upper-body manipulation, demonstrating superior whole-body coordination.

In terms of locomotion, even without perception, \pilot exhibits robust capability with minimal tracking errors for linear and angular velocity ($E_\mathbf{v}=0.145, E_\omega=0.099$). This performance advantage stems from our unified control architecture, which synergistically coordinates all joints regardless of the visual input. In contrast, baselines with decoupled controllers (HOMIE, FALCON) often sacrifice locomotion agility to mitigate disturbances from the upper body, resulting in significantly higher velocity errors ($E_\mathbf{v} > 0.270$).

Regarding torso control, HOMIE and FALCON suffer from limited workspace, lacking actuation for torso pitch and roll. While HOMIE exhibits the lowest yaw tracking error ($E_y=0.020$), this is an artifact of a stiff PD controller acting on a DoF that induces minimal Center of Mass (CoM) shifting. However, this rigidity severely compromises its adaptability, as evidenced by its poor locomotion performance. Conversely, both AMO and \pilot support full torso actuation. \pilot outperforms AMO in orientation tracking ($E_r, E_p, E_y$) and achieves substantial precision in base height tracking ($E_h=0.010$). We attribute this precision to the MoE-based policy, which effectively synthesizes diverse motion experts. Furthermore, enabled by the residual learning framework and holistic optimization, \pilot maintains robust upper-body manipulation ($E_{\mathbf{arm}}=0.206$), matching the best baseline while providing a significantly larger effective workspace compared to mocap-dependent methods like FALCON.

Overall, \pilot functions as a versatile low-level controller for loco-manipulation, delivering superior command tracking performance alongside an expansive operational workspace. It supports a wide vertical range of base heights ($h \in [0.3, 0.76]\,\text{m}$), enabling behaviors ranging from low-squatting to full standing. Furthermore, \pilot offers extensive torso control authority—including roll ($[-0.4, 0.4]\,\text{rad}$), pitch ($[-0.3, 1.5]\,\text{rad}$), and yaw ($[-1.5, 1.5]\,\text{rad}$)—to facilitate complex arm-torso coordination tasks that fully cover the kinematic reach of the upper limbs.
\subsubsection{\textbf{Ablation on policy learning}}

To address Q2 and evaluate the necessity of visual perception, we benchmark the full \pilot framework against three ablated variants (w/o vision, w/o attention-based encoder, and w/o MoE) on a diverse suite of challenging terrains, including staircases, elevated platforms, slopes, and randomized rough surfaces.
To quantitatively assess traversal safety, we introduce a foot stumble metric, $E_{\text{stumble}}$, defined as the frequency of collision events where horizontal foot impact forces exceed vertical support forces: $E_{\text{stumble}} = \frac{1}{T} \sum_{t=1}^{T} \mathbb{I}(|F_{\text{foot}}^{xy}| \ge F_{\text{foot}}^z)$, where $\mathbb{I}(\cdot)$ is the indicator function. A lower $E_{\text{stumble}}$ indicates smoother terrain interaction and higher stability during loco-manipulation.

As detailed in Table \ref{tab:errors_comparison}(b), the removal of key components leads to significant performance degradation. The w/o vision variant exhibits the most severe decline across all metrics, confirming that proprioception alone is fundamentally insufficient for robust loco-manipulation on complex terrains. Furthermore, even with visual input, the quality of representation proves critical: replacing the attention-based encoder with a simple MLP results in a dramatic deterioration in locomotion safety, with the stumble rate $E_{\text{stumble}}$ escalating by an order of magnitude (from 0.006 to 0.066).
This empirical evidence strongly supports our hypothesis that the attention mechanism is critical for identifying optimal steppable areas and extracting state-dependent geometric features, thereby facilitating efficient and safe traversal over unstructured terrains.

\begin{figure}[!t]
\centering
\includegraphics[width=0.95\linewidth]{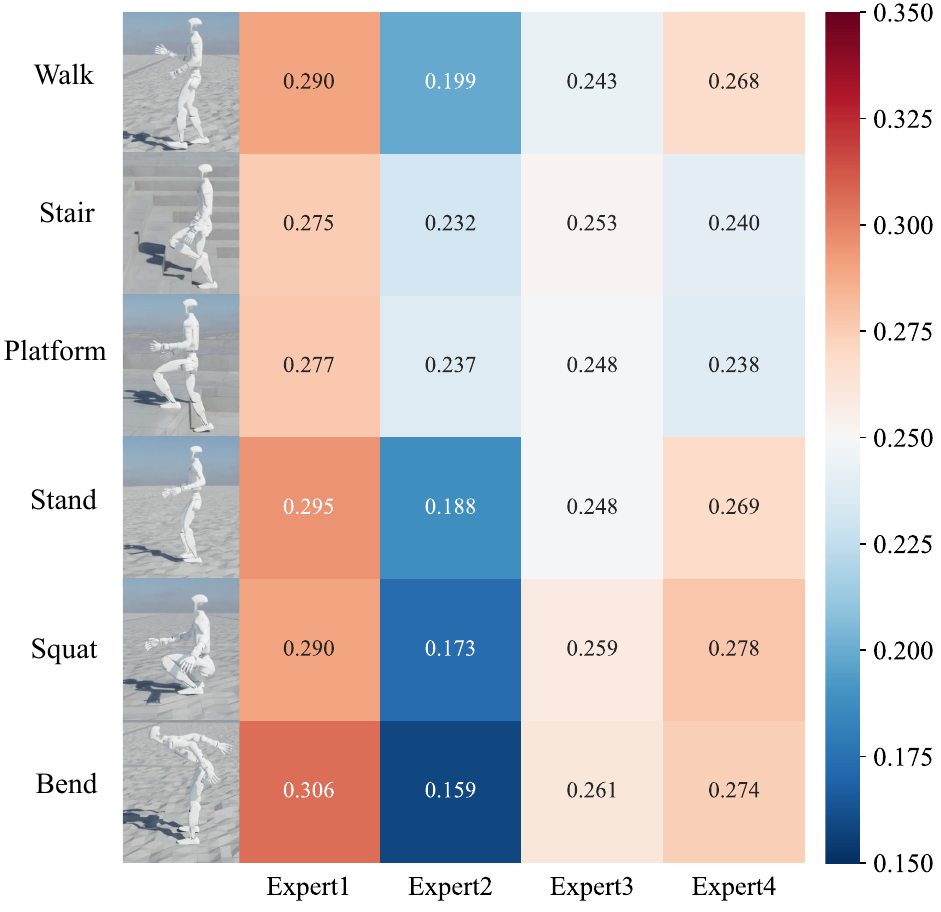}
\vspace{-10pt}
\caption{\textbf{Visualization of expert activation across six motion modes.} The color intensity represents the magnitude of the activation probability, where deep red denotes higher activation and deep blue indicates lower activation.}
\label{fig_heatmap}
\vspace{-18pt}
\end{figure}

Furthermore, replacing the unified MoE policy with a standard monolithic network (w/o MoE) significantly impairs the robot's ability to track velocities and maintain precise torso orientation. This result underscores the necessity of the MoE architecture, which leverages specialized experts to focus on distinct motion patterns. By dynamically coordinating these diverse skills, the MoE framework enables the robot to adaptively switch between locomotion and manipulation modes, effectively mitigating the gradient interference often observed in monolithic policies.

To interpret the learned MoE specialization, we visualize the average gating activation weights across six distinct motion primitives in Fig. \ref{fig_heatmap}. The heatmap reveals a clear pattern where tasks requiring similar skills elicit comparable expert activations. Specifically, for dynamic traversal tasks (e.g., traversing  stairs and elevated platforms) that necessitate accurate environmental perception and precise foot placement, Expert 2 exhibits a marked increase in activation (rising from $\approx 0.16$ in static tasks to $0.23$). Concurrently, the reliance on Experts 1 and 4 is reduced compared to stationary scenarios. In contrast, proprioception-dominant motions like squatting and deep bending demonstrate a strong preference for Expert 1 and Expert 4. For instance, in the deep bending maneuver, Expert 1 achieves a peak activation of $0.306$, while the influence of the perception-oriented Expert 2 diminishes significantly to $0.159$. This indicates a learned functional decoupling: Expert 2 has specialized in processing exteroceptive features for robust terrain interaction, whereas Experts 1 and 4 govern whole-body coordination and balance during large-workspace manipulations. These findings empirically validate the distinct specialization of our policy and its capability for adaptive skill switching based on real-time contextual demands.

\begin{figure*}[!t]
\centering
\includegraphics[width=0.98\linewidth]{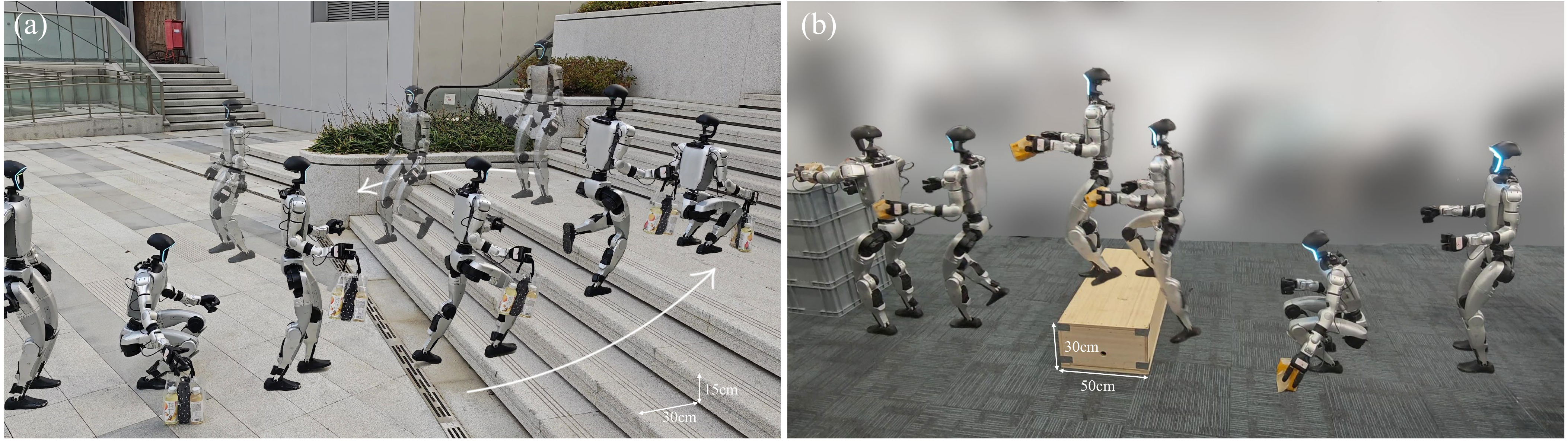}
\vspace{-10pt}
\caption{\textbf{Real-world Experiments.} \pilot successfully executes object transport tasks across challenging terrains. The robot is shown traversing (a) a staircase and (b) a high platform while carrying a payload.}

\label{fig_mainexp}
\vspace{-18pt}
\end{figure*}

\vspace{-10pt}
\subsection{Real-world Results}

To validate the efficacy and robustness of \pilot in real-world scenarios, we directly deploy the policy trained in simulation onto the physical Unitree G1 humanoid robot via zero-shot transfer. The policy network infers actions at $50\,\text{Hz}$, which are subsequently transmitted to the low-level motor PD controllers executing at $500\,\text{Hz}$. For environmental perception, we implemented a LiDAR-based robot-centric elevation mapping system, inspired by the framework in \cite{PIM}. This perception pipeline leverages Fast-LIO \cite{fastlio} for state estimation and probabilistic terrain mapping \cite{elevation1} to generate local height maps in real-time. The processed terrain information is published to the policy network at an update rate of $10\,\text{Hz}$. 

\subsubsection{\textbf{Teleoperation Results}} Fig. \ref{fig_mainexp} illustrates a sequence of challenging loco-manipulation tasks using VR teleoperation performed in complex 3D environments, highlighting the system's capability for whole-body coordination and robust terrain traversal.

\textbf{Object Transport via Stair Traversal:} As depicted in Fig. \ref{fig_mainexp}(a), this challenging long-horizon task necessitates the execution of four critical phases: (1) deep squatting to retrieve a weighted bag (containing two water bottles, total mass $\approx 1.0\,\text{kg}$) from the ground, which demands coordinated arm-torso movements; (2) transitioning to a standing posture and traversing a multi-step staircase while carrying the load; (3) squatting to place the payload at a designated target; and (4) standing, turning, and descending the stairs.

This scenario serves as a rigorous stress test for the synergistic integration of perceptive locomotion and stable manipulation. The complexity lies in the bidirectional coupling between mobility and manipulation:
\begin{itemize}
    \item \textbf{Locomotion-Induced Instability}: The robot must execute precise foot placement on the stair treads to prevent stumbling. Any lower-body instability would instantaneously propagate to the base, severely compromising the precision of the upper-body end-effector.
    \item \textbf{Manipulation-Induced Disturbances}: Simultaneously, the asymmetric load at the end-effectors introduces a dynamic disturbance that shifts the Center of Mass (CoM). The controller must perform robust force adaptation and precise arm tracking to ensure that the manipulation task does not destabilize the locomotion gait, particularly when traversing uneven terrain.
\end{itemize}

Fig. \ref{fig_mainexp}(a) demonstrates that \pilot successfully handles these coupled dynamics. Across five trials with varying payloads, the robot achieves a $100\%$ success rate ($5/5$) in executing the object transport task via stair traversal. Throughout the maneuver, the robot maintains a stable and stumble-free gait while keeping the payload steady, validating \pilot's robustness and adaptability as a perceptive, integrated whole-body controller.

\textbf{Object Transport via High Platform Traversal}: Analogous to the previous scenario, this task entails a multi-stage sequence: (1) performing a deep squat and torso flexion to retrieve a tissue box from the ground; (2) traversing an elevated platform while maintaining a stable grasp; and (3) placing the object onto a target container.
We emphasize that the geometric constraints of the platform—specifically its height of $0.3\,\text{m}$ and limited width of $0.5\,\text{m}$—introduce substantial difficulties. For a mid-sized humanoid robot, a $0.3\,\text{m}$ vertical step represents a severe kinematic challenge, requiring a vertical clearance exceeding $40\%$ of the robot's leg length, thereby necessitating near-limit joint articulation and high-torque generation. 
Consequently, this imposes stringent foot placement accuracy to avoid edge collisions or misstepping, a challenge further compounded by the need for upper-body stability during transport.
As demonstrated in Fig. \ref{fig_mainexp}(b), \pilot exhibits exceptional whole-body coordination and seamless skill transitions throughout the maneuver. Crucially, the policy maintains a stable object grasp even during the dynamic locomotion phase over the elevated terrain, further validating  the robustness of our unified control architecture.

\begin{figure}[!t]
\centering
\includegraphics[width=0.95\linewidth]{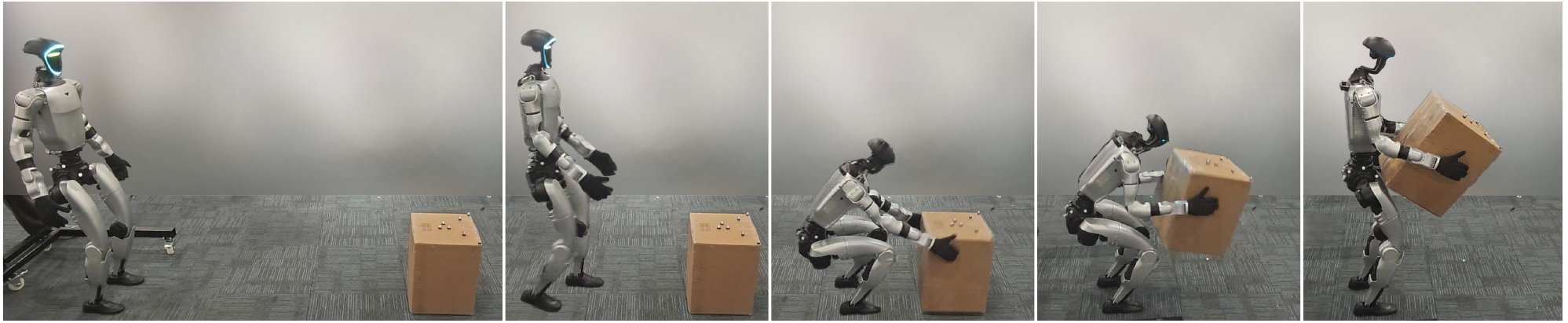}
\vspace{-10pt}
\caption{\textbf{Autonomous policy execution.} With \pilot serving as the low-level tracking controller, the high-level policy directs the robot to approach the target box, grasp and lift the payload, and recover to an upright configuration.}
\label{fig_liftbox}
\vspace{-18pt}
\end{figure}
\subsubsection{\textbf{Autonomous Task via Hierarchical Integration}}
To further validate the versatility and robustness of \pilot as a foundational low-level controller, we develop a hierarchical control framework to execute a \textit{Lift-Box} task autonomously without teleoperation. Specifically, we train a high-level policy designed to function as a planner. This policy operates on an observation space defined as $o_t^{\text{high}} = [p_{o_t}^{\text{rel}}, R_{o_t}^{\text{rel}}, \omega_t^{\text{base}}, g_t, q_t, \dot{q}_t, c_{t-1}^{\text{high}}]$, where $p_{o_t}^{\text{rel}}$ and $R_{o_t}^{\text{rel}}$ denote the relative position and orientation of the target object with respect to the robot's base frame. Based on these inputs, the high-level policy generates the requisite whole-body commands $c_t^{\text{high}} = [v_t^x, v_t^y, \omega_t^{\text{yaw}}, h_t^{\text{base}}, \mathbf{rpy}_t, q_t^{\text{upper}*}]$, which are transmitted to \pilot for execution.
The high-level policy is trained using RL in simulation and transferred to the physical robot via zero-shot transfer. For real-world state estimation, we attach mocap markers to the robot's pelvis and the target box, enabling real-time acquisition of their relative poses.

As depicted in Fig. \ref{fig_liftbox}, the robot successfully executed the full autonomous sequence: navigating to the target, performing a deep squat with torso flexion, bimanually grasping the box, and recovering to a stable standing posture. These results empirically demonstrate that \pilot serves as a stable and robust low-level controller that effectively abstracts whole-body dynamics, allowing for seamless integration with high-level planners to achieve coordinated locomotion and manipulation.

\section{Conclusion}
In this paper, we propose \pilot, a unified single-stage RL framework tailored for humanoid loco-manipulation on complex terrains, which seamlessly integrates perceptive locomotion with expansive whole-body control. 
Through extensive experiments in both simulation and real-world scenarios, we demonstrated that \pilot significantly outperforms prior low-level controllers in terms of command tracking precision, terrain traversability, and system robustness. These findings empirically validate the effectiveness of our approach. We believe that \pilot establishes a solid foundation for practical, stable loco-manipulation systems capable of operating in complex scenarios beyond simple flat ground.
In future work, we aim to develop a high-level visuomotor policy that utilizes \pilot as a robust low-level controller to autonomously execute long-horizon loco-manipulation tasks in unstructured environments, thereby eliminating the reliance on teleoperation or external motion capture systems.


\bibliographystyle{IEEEtrans}  
\bibliography{references} 

\end{document}